\documentclass[conference]{IEEEtran}
\usepackage{blindtext, graphicx}
\usepackage[backend=bibtex,style=ieee,sorting=ynt]{biblatex}
\usepackage{pgfplots}   
\usepackage{SIunits}     
\usepackage{amsmath}

\begin{document}
\title{Learning Approximate Neural Estimators for Wireless Channel State Information}

\author{
\IEEEauthorblockN{Tim O'Shea}
\IEEEauthorblockA{Electrical and Computer Engineering\\
Virginia Tech, Arlington, VA\\
oshea@vt.edu}
\and
\IEEEauthorblockN{Kiran Karra}
\IEEEauthorblockA{Electrical and Computer Engineering\\
Virginia Tech, Arlington, VA\\
kiran.karra@vt.edu}
\and
\IEEEauthorblockN{T. Charles Clancy}
\IEEEauthorblockA{Electrical and Computer Engineering\\
Virginia Tech, Arlington, VA\\
tcc@vt.edu}
}

\maketitle

\begin{abstract}
Estimation is a critical component of synchronization in wireless and signal processing systems.  There is a rich body of work on estimator derivation, optimization, and statistical characterization from analytic system models which are used pervasively today.  We explore an alternative approach to building estimators which relies principally on approximate regression using large datasets and large computationally efficient artificial neural network models capable of learning non-linear function mappings which provide compact and accurate estimates.  For single carrier PSK modulation, we explore the accuracy and computational complexity of such estimators compared with the current gold-standard analytically derived alternatives.  We compare performance in various wireless operating conditions and consider the trade offs between the two different classes of systems.  Our results show the learned estimators can provide improvements in areas such as short-time estimation and estimation under non-trivial real world channel conditions such as fading or other non-linear hardware or propagation effects.
\end{abstract}

\IEEEpeerreviewmaketitle

\section{Introduction}
Estimation is a critical component of many signal processing systems.  We focus in this paper on several key synchronization tasks which are used widely in the reception of single carrier modulated phase-shift keying (PSK) signals.  Specifically, we look at carrier frequency offset (CFO) estimation and timing estimation which are used to recover the time and frequency of a received signal transmission to enable demodulation and information recovery.  Excellent estimators exist for these tasks and are derived analytically from expressions for the signal model and the channel model using techniques such as maximum-likelihood (MLE) or minimum-mean squared error (MMSE) optimization metrics to produce an expression for an estimator.  This approach works well, and has worked in many communications systems over the year; however it suffers from several shortcomings:

\begin{enumerate}
    \item It relies on an accurate analytic model of the signal and channel, the signal is man-made and typically easy to model, but hardware defects and distortion effects along with channel propagation effects or distributions in specific environments are rarely captured in great detail when performing this optimization.  Analytic estimators derived under Gaussian channel assumptions are often used by default.
    \item It requires a manual analysis process to derive a well conditioned estimator for a specific signal type, modulation type, and/or set of reference tones which is time consuming and often leads to time and cost in system engineering.
    \item Algorithms formed through the analytic process are typically kept compact (minimum number of terms) in order to facilitate the ease of analysis.  However, from a power consumption perspective, such a compact form may not be the most concurrent form of an estimator, where algorithmic concurrency often leads to the lowest power algorithm on mobile platforms.
\end{enumerate}

Artificial neural networks (ANNs) have been used for regression tasks previously \cite{specht1991general, cochocki1993neural}, many of these involving signal processing tasks or transformations.  However numerous advances have been made in ANNs over the past few years including significant improvements in gradient descent, regularization, network architecture and activation functions, as well as the use of many-core compute platforms such as graphics processing units (GPUs) to realize and train very large networks rapidly.  These advances have led to the growth of 'deep learning', where training very large neural networks which were previously in-feasible (for instance in the 90s when many of the prior works in this area were published) can now be accomplished easily with commercial hardware and open source software such as TensorFlow \cite{tensorlow} and Keras \cite{keras}.  Recent efforts \cite{DBLP:journals/corr/OSheaH17} have shown this class of deep learning for physical layer representations and algorithms to be effective and competitive with modern baselines.

The outline of our paper is as follows.  We begin by discussing expert estimator approaches to estimating the timing and frequency offsets for PSK burst signals.  We then describe a new, deep learning based approach for estimating these quantities.  After estimation is discussed, we conduct experiments with both of these approaches and compare the performance.  Finally, concluding remarks are provided.

\section{Background - Expert Estimation}
In this section, we introduce and describe the expert features based center frequency offset estimation and timing offset estimation techniques.  

\subsection{Center Frequency Offset Estimation}
A common approach to center frequency offset estimation is to use an FFT based technique which estimates the frequency offset by using a periodogram of the $m^{th}$ power of the received signal \cite{wang2004non}.  The frequency offset detected by this technique is then given by:

\begin{align}\label{eq:fft_cfo}
    \Delta \hat{f} = \frac{F_s}{N \cdot m} \underset{f}{\mathrm{argmax}}    \lvert \sum_{k=0}^{N-1} r^m[k] e^{-j 2 \pi k t / N} \rvert \\
    \left( -\frac{R_{sym}}{2} \leq f \leq \frac{R_{sym}}{2} \right) \nonumber,
\end{align}

\begin{figure}[h]
    \centering
    \includegraphics[width=0.5\textwidth]{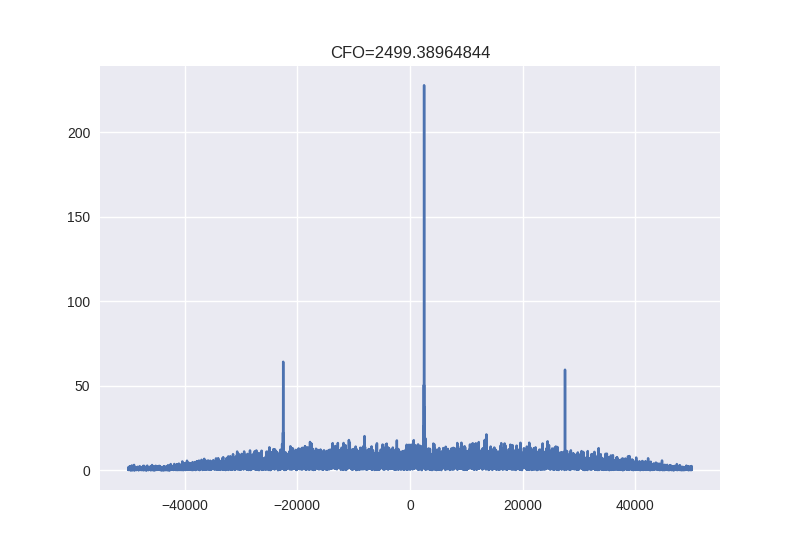}
    \caption{CFO Expert Estimator Power Spectrum with simulated 2500 Hz offset}
    \label{fig:cfo_expert_explain}
\end{figure}

where $m$ is the modulation order, $r(k)$ is the received sequence, $R_{sym}$ is the symbol rate, $F_s$ is the sampling frequency, and $N$ is the number of samples.  The algorithm searches for a frequency that maximizes the time average of the $m^{th}$ power of the received signal over various frequencies in the range of $\left( -\frac{R_{sym}}{2} \leq f \leq \frac{R_{sym}}{2} \right)$.  Due to the algorithm operating in the frequency domain, the center frequency offset manifests as the maximum peak in the spectrum of $r^m(k)$.  Fig. \ref{fig:cfo_expert_explain} shows an example cyclic spectrum for a QPSK signal with a 2500 Hz center frequency offset, where the peak indicates the center frequency offset for the burst.

\subsection{Timing Offset Estimation}
In traditional wireless communications systems, timing offset estimation is performed by matched filtering the received sequence to a known preamble sequence.  The time-offset which maximizes the output of the matched filter's convolution is then taken to be the time-offset of the received signal.  Matched filtering can be represented by (\ref{eq:mf})

\begin{equation} \label{eq:mf}
y[k] = \sum_{k=-\infty}^{k=\infty} h[n-k] r[k],
\end{equation}
where $h[k]$ is the preamble sequence.  The matched-filter is known as the optimal filter for maximizing the signal to noise ratio (SNR) in the presence of additive stochastic white noise.  

\section{Technical Approach for Learned Algorithms}
In this section, we introduce our approach for learning algorithms to estimate the CFO and timing offsets.  

Our approach to learned estimator generation relies on construction, training and evaluating an ANN based on a representative dataset.  When relying on learned estimators, much of work and difficulty lies in generating a dataset which accurately reflects the final usage conditions desired for the estimator.  In our case, we produce numerous examples of wireless emissions in complex baseband sampling with rich channel impairment effects which are designed to match the intended real world conditions the system will operate in.  We associate target labels from ground truth for center frequency offset and timing error which are used to optimize the estimator.  

To train an ANN model, we consider the minimization of mean-squared error (MSE) and log-cosine hyperbolic (log-cosh) \cite{catoni2012challenging} and Huber loss functions (shown in table \ref{tab:losses}).  The latter are known to have improved properties in robust learning, which may benefit such a regression learning task on some datasets and tasks.  In our initial experiments in this paper, we observe the best quantitative performance using the MSE loss function which we shall use for the remainder.

\begin{table}
\renewcommand{\arraystretch}{1.0}
\centering
\caption{Common Regression Loss Functions}
\label{tab:losses}
\begin{tabular}{l|c}
Method & Expression \\
\hline
MSE & $L_{MSE}(y,\hat{y}) =  \sum_i \left( y_i - \hat{y_i} \right)^2 $\\
MAE & $L_{MAE}(y,\hat{y}) =  \sum_i abs \left( y_i - \hat{y_i} \right) $\\
log-cosh & $L_{LogCosh}(y,\hat{y}) =  \sum_i log \left( cosh \left( y_i - \hat{y_i} \right) \right) $\\
Huber & $L_{Huber}(y,\hat{y}) = \sum_i $ 
$ \begin{cases} 
      \frac{1}{2} \left( y_i - \hat{y_i} \right)^2   & abs(y_i-\hat{y_i}) < 1 \\
      \left( y_i - \hat{y_i} \right)  & abs(y_i-\hat{y_i}) \geq 1 
      \end{cases}
$ \\ 

\hline
\end{tabular}
\end{table}

We search over a large range of model architectures using Adam \cite{adam} to perform gradient descent to optimize for model parameters based on our training dataset.  Ultimately this optimization and model search may be a trade off where the search is for a model of a minimal or limited complexity which achieves a satisfactory level of performance.  However in this paper we do not limit or optimize for model complexity, choosing only the best performing model based on MSE.

ANN architectures used for our performance evaluation are shown below, both stacked convolutional neural networks with narrowing dimensions which map noisy wide initial time series data down to a compact single valued regression output.  In the case of CFO estimation architecture shown in Table \ref{tab:cfonn}, we find that an average pooling layer works well to help improve performance and generalization of the initial layer feature maps, while in the timing estimation architecture in table \ref{tab:timenn} no-pooling, or max-pooling tends to work better.  This makes sense on an intuitive level as CFO is distilling all symbols received throughout the input into a best frequency estimate, while timing in a traditional matched filter sense, is derived typically from a maximum response at a single offset.  

\begin{table}
\renewcommand{\arraystretch}{1.0}
\centering
\caption{ANN Architecture Used for CFO Estimation}
\label{tab:cfonn}
\begin{tabular}{l|c}
 Layer    & Output dimensions    \\\hline
 Input & (nsamp,2) \\
 Conv1D + ReLU & (variable,32) \\
 AveragePooling1D & (variable,32) \\
 Conv1D + ReLU & (variable,128) \\
 Conv1D + ReLU & (variable,256) \\
 Linear & $1$
\end{tabular}
\end{table}

\begin{table}
\renewcommand{\arraystretch}{1.0}
\centering
\caption{ANN Architecture Used for Timing Estimation}
\label{tab:timenn}
\begin{tabular}{l|c}
 Layer    & Output dimensions    \\\hline
 Input & (2048,2) \\
 Conv1D + ReLU & (511,32) \\
 Conv1D + ReLU & (126,64) \\
 Conv1D + ReLU & (30,128) \\
 Conv1D + ReLU & (2,256) \\
 Dense + Linear & (1)
\end{tabular}
\end{table}

\section{Performance Analysis}
In this section, we discuss the methodology used to evaluate expert estimators with learn estimators.  We begin by discussing how the datasets used for evaluating the performance between these approaches are generated.  We then discuss the test methodology.  Finally, we discuss the results of the different approaches.  

\subsection{Data Generation}

\begin{figure}[h]
    \centering
    \includegraphics[width=0.5\textwidth]{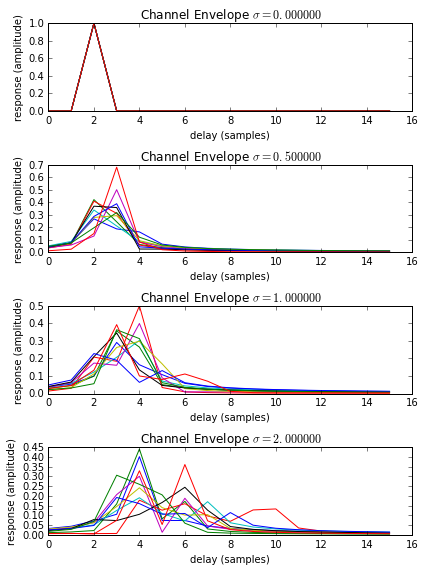}
    \caption{Impulse Response Modes Compared}
    \label{fig:impulse}
\end{figure}

We generate two different sets of data for evaluating the performance of the two competing approaches.  All generated data are based off of QPSK bursts with random IID symbols, and shaped with a RRC-filter with a roll-off $\beta=0.25$ and a filter span of 6, and sampled at 400 kHz with a symbol rate of 100 kHz.  We consider 4 channel conditions, additive white Gaussian noise (AWGN) with no fading, and three cases of Rayleigh fading with varying mean delay spreads in samples of $\sigma=0.5,1,2$.  Amplitude envelopes for a number of complex valued channel responses for each of these delay spreads are shown in figure \ref{fig:impulse} to provide some visual insight into the impact of Rayleigh fading effects at each of these delays.  For the last case, significant inter-symbol interference (ISI) is present in the data.

The first dataset generated is the timing dataset, in which we prepend the burst with a known preamble of 64 symbols and random noise samples at the same SNR as the data portion of the burst.  The number of noise samples preprended is drawn from a $\mathcal{U} \sim (0,1.25)$, in units of milliseconds.  Additionally, a random phase offset drawn from a $\mathcal{U} \sim (0,2 \pi)$ is introduced for each burst in the dataset.

The second dataset generated is the center frequency offset data, in which every example burst has a center frequency offset drawn from a $\mathcal{U} \sim (-50e3,50e3)$ distribution, in units of Hz.  The bounds of this correspond to half the symbol rate, $R_{sym}/2$.  Additionally, a random phase offset drawn from a $\mathcal{U} \sim (0,2 \pi)$ is introduced for each burst in the dataset.

These datasets are generated for SNR's of 0 dB, 5 dB, and 10 dB and for an AWGN channel and three different Rayleigh fading channels with different mean delay spread values (0.5, 1, and 2) representing different levels of reflection in a given wireless channel environment.  We store the label of the timing offset and center frequency offsets as ground truth for training and evaluation.

\subsection{Test Methodology and Experimental Results}

For each dataset generated above we optimize network weights using Adam for 100 epochs, reducing the initial learning rate of $1e-3$ by a factor of two for each 10 epochs with no reduction in validation loss, ultimately using the parameters corresponding to the epoch with the lowest validation loss.   With the datasets generated above, we then compute the test error using a separate data partition between ground truth labels for timing and center frequency offset and predicted values generated using both expert and deep learning/ANN based estimators.  The standard deviation of the estimator error is used as our metric for comparison.

\subsubsection{Timing Results}

\begin{figure}[h]
    \centering
    \includegraphics[width=0.5\textwidth]{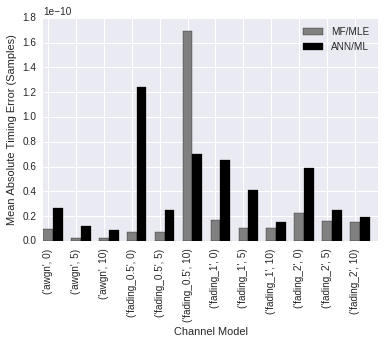}
    \caption{Timing Estimation Mean Error Comparison }
    \label{fig:timeR}
\end{figure}

In the timing estimation comparison, we show estimator mean-absolute-error results in figure \ref{fig:timeR}.  Inspecting these results we can see that the traditional matched filtering maximum likelihood approach (MF/MLE) achieves excellent performance under the AWGN channel condition.  We can see significant degradation of the MF/MLE baseline accuracy under the fading channel models however as a simple matched filter MLE timing estimation approach has no ability to compensate for the expected range of channel delay spreads.  In this case the artificial neural network / machine learning (ANN/ML) estimator approach on average can not attain equivalent performance in all or even most cases.  However, we see that this approach does attain an average absolute error within the same order of magnitude, and does in some fading cases achieve a lower  mean absolute error in the case of a fading channel.  

\subsubsection{Center Frequency Results}

\begin{figure}[h]
\centering
\begin{tikzpicture}
\begin{axis}[
	xmode=log,
	ymode=log,
	xlabel=Block Size (samples), 
	ylabel=Estimator Std. Deviation (Hz),
	title={CFO Estimation in AWGN Channel},
	grid=both,
	minor grid style={gray!25},
	major grid style={gray!25},
	width=0.85\linewidth,
    legend style={font=\fontsize{5}{6}\selectfont},
    legend pos=south west,
	no marks]
\addplot[line width=1pt,solid,color=red] %
	table[x=len,y=ml,col sep=comma]{cfo_awgn_0.csv};
\addlegendentry{ML Estimator 0dB};
\addplot[line width=1pt,solid,color=blue] %
	table[x=len,y=ml,col sep=comma]{cfo_awgn_5.csv};
\addlegendentry{ML Estimator 5dB};
\addplot[line width=1pt,solid,color=green] %
	table[x=len,y=ml,col sep=comma]{cfo_awgn_10.csv};
\addlegendentry{ML Estimator 10dB};
\addplot[line width=1pt,dashed,color=red] %
	table[x=len,y=expert,col sep=comma]{cfo_awgn_0.csv};
\addlegendentry{MAP Estimator 0dB};
\addplot[line width=1pt,dashed,color=blue] %
	table[x=len,y=expert,col sep=comma]{cfo_awgn_5.csv};
\addlegendentry{MAP Estimator 5dB};
\addplot[line width=1pt,dashed,color=green] %
	table[x=len,y=expert,col sep=comma]{cfo_awgn_10.csv};
\addlegendentry{MAP Estimator 10dB};
\end{axis}
\end{tikzpicture}
\caption{Mean CFO Estimation Error for AWGN Channel}
\label{fig:r1}
\end{figure}
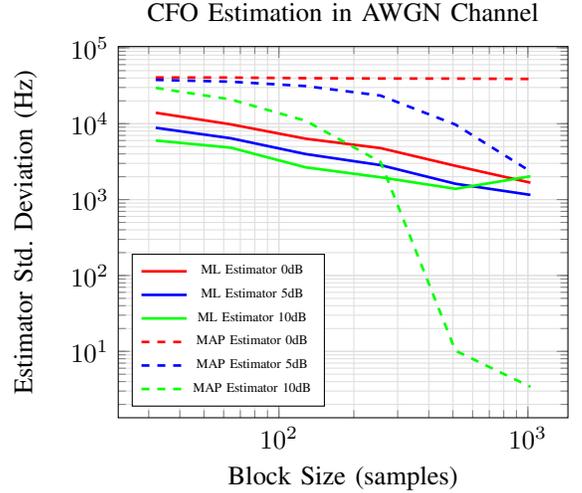

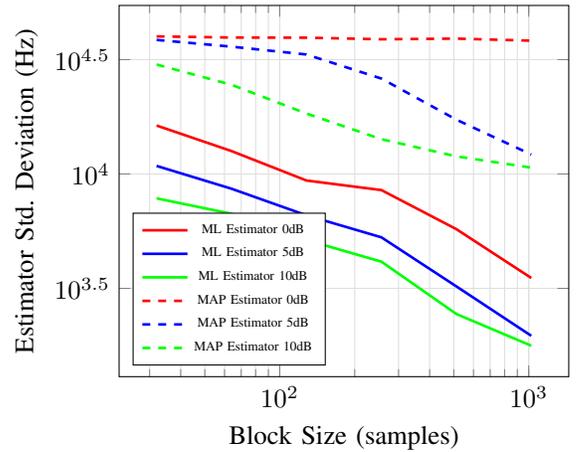
\begin{figure}[h]
\centering
\begin{tikzpicture}
\begin{axis}[
	xmode=log,
	ymode=log,
	xlabel=Block Size (samples), 
	ylabel=Estimator Std. Deviation (Hz),
	title={CFO Estimation in Light Rayleigh Channel},
	grid=both,
	minor grid style={gray!25},
	major grid style={gray!25},
	width=0.85\linewidth,
    legend style={font=\fontsize{5}{6}\selectfont},
    legend pos=south west,
	no marks]
\addplot[line width=1pt,solid,color=red] %
	table[x=len,y=ml,col sep=comma]{cfo_fading_0.5_0.csv};
\addlegendentry{ML Estimator 0dB};
\addplot[line width=1pt,solid,color=blue] %
	table[x=len,y=ml,col sep=comma]{cfo_fading_0.5_5.csv};
\addlegendentry{ML Estimator 5dB};
\addplot[line width=1pt,solid,color=green] %
	table[x=len,y=ml,col sep=comma]{cfo_fading_0.5_10.csv};
\addlegendentry{ML Estimator 10dB};
\addplot[line width=1pt,dashed,color=red] %
	table[x=len,y=expert,col sep=comma]{cfo_fading_0.5_0.csv};
\addlegendentry{MAP Estimator 0dB};
\addplot[line width=1pt,dashed,color=blue] %
	table[x=len,y=expert,col sep=comma]{cfo_fading_0.5_5.csv};
\addlegendentry{MAP Estimator 5dB};
\addplot[line width=1pt,dashed,color=green] %
	table[x=len,y=expert,col sep=comma]{cfo_fading_0.5_10.csv};
\addlegendentry{MAP Estimator 10dB};
\end{axis}
\end{tikzpicture}
\caption{Mean CFO Estimation Error (Fading $\sigma$=0.5)}
\label{fig:r2}
\end{figure}

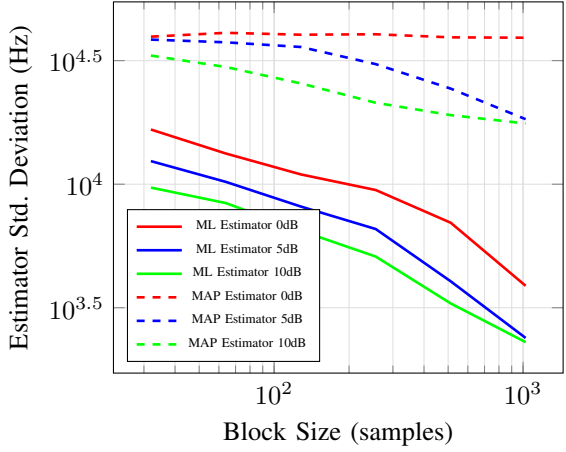
\begin{figure}[h]
\centering
\begin{tikzpicture}
\begin{axis}[
	xmode=log,
	ymode=log,
	xlabel=Block Size (samples), 
	ylabel=Estimator Std. Deviation (Hz),
	title={CFO Estimation in Medium Rayleigh Channel},
	grid=both,
	minor grid style={gray!25},
	major grid style={gray!25},
	width=0.85\linewidth,
    legend style={font=\fontsize{5}{6}\selectfont},
    legend pos=south west,
	no marks]
\addplot[line width=1pt,solid,color=red] %
	table[x=len,y=ml,col sep=comma]{cfo_fading_1_0.csv};
\addlegendentry{ML Estimator 0dB};
\addplot[line width=1pt,solid,color=blue] %
	table[x=len,y=ml,col sep=comma]{cfo_fading_1_5.csv};
\addlegendentry{ML Estimator 5dB};
\addplot[line width=1pt,solid,color=green] %
	table[x=len,y=ml,col sep=comma]{cfo_fading_1_10.csv};
\addlegendentry{ML Estimator 10dB};
\addplot[line width=1pt,dashed,color=red] %
	table[x=len,y=expert,col sep=comma]{cfo_fading_1_0.csv};
\addlegendentry{MAP Estimator 0dB};
\addplot[line width=1pt,dashed,color=blue] %
	table[x=len,y=expert,col sep=comma]{cfo_fading_1_5.csv};
\addlegendentry{MAP Estimator 5dB};
\addplot[line width=1pt,dashed,color=green] %
	table[x=len,y=expert,col sep=comma]{cfo_fading_1_10.csv};
\addlegendentry{MAP Estimator 10dB};
\end{axis}
\end{tikzpicture}
\caption{Mean CFO Estimation Error (Fading $\sigma$=1)}
\label{fig:r3}
\end{figure}

\begin{figure}[h]
\centering
\begin{tikzpicture}
\begin{axis}[
	xmode=log,
	ymode=log,
	xlabel=Block Size (samples), 
	ylabel=Estimator Std. Deviation (Hz),
	title={CFO Estimation in Harsh Rayleigh Channel},
	grid=both,
	minor grid style={gray!25},
	major grid style={gray!25},
	width=0.85\linewidth,
    legend style={font=\fontsize{5}{6}\selectfont},
    legend pos=south west,
	no marks]
\addplot[line width=1pt,solid,color=red] %
	table[x=len,y=ml,col sep=comma]{cfo_fading_2_0.csv};
\addlegendentry{ML Estimator 0dB};
\addplot[line width=1pt,solid,color=blue] %
	table[x=len,y=ml,col sep=comma]{cfo_fading_2_5.csv};
\addlegendentry{ML Estimator 5dB};
\addplot[line width=1pt,solid,color=green] %
	table[x=len,y=ml,col sep=comma]{cfo_fading_2_10.csv};
\addlegendentry{ML Estimator 10dB};
\addplot[line width=1pt,dashed,color=red] %
	table[x=len,y=expert,col sep=comma]{cfo_fading_2_0.csv};
\addlegendentry{MAP Estimator 0dB};
\addplot[line width=1pt,dashed,color=blue] %
	table[x=len,y=expert,col sep=comma]{cfo_fading_2_5.csv};
\addlegendentry{MAP Estimator 5dB};
\addplot[line width=1pt,dashed,color=green] %
	table[x=len,y=expert,col sep=comma]{cfo_fading_2_10.csv};
\addlegendentry{MAP Estimator 10dB};
\end{axis}
\end{tikzpicture}
\caption{Mean CFO Estimation Error (Fading $\sigma$=2)}
\label{fig:r4}
\end{figure}
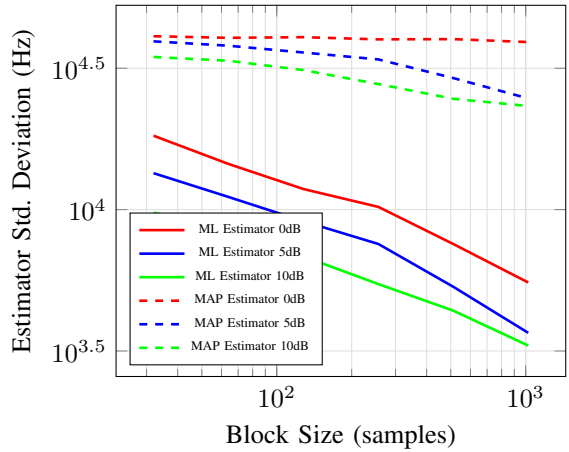

Quantitative results for estimation of center frequency offset error are shown in figures  \ref{fig:r1},\ref{fig:r2},\ref{fig:r3},\ref{fig:r4}, summarizing the performance of both the baseline maximum likelihood (MLE) method with dashed lines and the ANN/ML method with solid lines.  We compare the mean absolute center frequency estimate error for each method at a range of different estimator block input length sizes.  As moment based methods generally improve for longer block sizes, we compare performance over a range of short-time examples to longer-time examples.  

In the AWGN case, in figure \ref{fig:r1} we can see that for 5 and 10dB SNR cases, by the time we reach a block size of 1024 samples, the baseline estimator is doing quite well, and for larger block sizes (above 1024 samples) with SNR of at least 5dB, performance of the baseline method is generally better.  However, even in the AWGN case, for small block sizes we are able to achieve lower error using the ANN/ML approach, even at low SNR levels of near 0dB.  

In the cases of fading channels shown in figures \ref{fig:r2},\ref{fig:r3},\ref{fig:r4}, we can see that performance of the baseline estimator degrades enormously from the AWGN case under which it was derived when delay spread is introduced.  Performance gets perpetually worse as $\sigma$ increases from 0.5 to 2 samples of mean delay spread.  In the case of the ANN/ML estimator, we also see a degradation of estimator accuracy as delay spread increases, but the effect is not nearly as dramatic, ranging from 3.4 to 23254 Hz in the MLE case (almost a ~7000x increase in error) versus a range of 2027 to 3305 Hz in the ML case (around a 1.6x increase in error).

\subsection{Computational Complexity}
In this section, we discuss the computational complexity of the expert estimators and the neural network based estimators.  Table \ref{tab:cfoflops} compares the approximate number of floating point operations (FLOPs) required to compute the center frequency offset estimate for both the expert estimator described above, and the neural network based estimator.  The FLOP counts for the expert estimator were derived by using the FFTW estimate for the number of flops required to compute an FFT, which scales with the the FFT size, $N$, as $5N log_2(N)$.  The FLOP counts reported are a function of the desired CFO frequency estimation resolution (1 Hz), the sampling frequency (400 kHz), and the input sequence length (parameterized in the table).  It is noticed in Table \ref{tab:cfoflops} that the expert estimator FLOP counts do not change with an increasing input sequence length.  This is because the FFT size required is, in this case, dominated by the desired frequency resolution of 1 Hz, rather than the input sequence length in this particular scenario.

The complexity of each layer of the neural network may be computed by considering the number of adds and multiplies within each layer.  These are given in terms of number of filter length $L$, input channels $ch_i$, output channels $ch_o$, output width $K$, input size $N_i$ and output size $N_o$ and pool size $p$.  Table \ref{tab:laycomp} enumerates the complexity of each layer type in terms of multiplies and adds as a function of layer parameters.

\begin{table}
\renewcommand{\arraystretch}{1.0}
\centering
\caption{Approx. Layer Complexity}
\label{tab:laycomp}
\begin{tabular}{c|c|c}
 Layer/Op    & Expression    \\\hline
 Conv1D/Mul  & $L*ch_i*ch_o*K$ \\
 Conv1D/Add  & $L*(ch_i+1)*ch_o*K$ \\
 Dense/Mul   & $N_i*N_o $ \\
 Dense/Add   & $(N_i+1)*N_o $ \\
 AvgPool/Add & $N_o * p $
\end{tabular}
\end{table}

In the case of the ANN complexity measurement, we observe that for small networks we are able to obtain lower complexity (rather than taking a high-resolution FFT).  While for the larger example sizes we perform a worst case of 5-10x more operations per estimation.  These results are not at all bad considering the ANN architectures were not optimized to minimize complexity.  When observing the complexity per layer for instance, it is clear that certain layers (for instance with large filter sizes) dominate the operation count, where they could likely be easily alleviated with smaller filter sizes and to obtain similar performance.  Effects of architecture optimization and reduced data representation precision will likely yield a complexity reduction on the order of 10-100x from these floating point numbers, providing an extremely competitive low-complexity estimator approximation with the baseline.

Similarly, Table \ref{tab:timingflops} shows the approximate number of floating point operations required to estimate the timing offset by both the expert estimators and the neural network based estimators.  The bulk of the FLOPs required for the estimation using the expert estimator depend on the input sequence length, because a matched filter must correlate across the entire sequence to find the optimal starting point.  

\begin{table}
\renewcommand{\arraystretch}{1.0}
\centering
\caption{CFO FLOP Count}
\label{tab:cfoflops}
\begin{tabular}{c|c|c}
 Sample Size    & Expert Estimator (MFlop)  & NN Estimator (MFlop)    \\\hline
 32   & 5.374 & 3.01 \\
 64   & 5.374 & 2.89 \\
 128  & 5.374 & 4.36 \\
 256  & 5.374 & 6.92 \\
 512  & 5.374 & 12.59 \\
 1024 & 5.374 & 23.71
\end{tabular}
\end{table}

\begin{table}
\renewcommand{\arraystretch}{1.0}
\centering
\caption{Timing FLOP Count}
\label{tab:timingflops}
\begin{tabular}{c|c|c}
 Sample Size    & Expert Estimator (MFlop)  & NN Estimator (MFlop)    \\\hline
 1024 & 1.05165 & 9.35
\end{tabular}
\end{table}

\section{Conclusion}

Estimation has been a fundamental building block of signal processing and wireless communications systems since their inception.  Optimization of analytically derived and statistically well-formed estimators has long been the building block upon which such systems have always been built, but is a non-trivially difficult process under channel and signal models of high complexity.  Our results are mixed, showing that in some cases existing benchmark estimators such as the matched filter are quite hard to beat using learned methods, but results in the same order of magnitude can be obtained readily given sufficient good labeled data and limited signal knowledge (for instance the ML timing estimator had no knowledge of the preamble used, the modulation type, the pulse shaping filter, etc) which potentially offers a lower implementation complexity if the cost of obtaining good labeled data is lower than that of imparting all the known reference signal information into the expert estimator.  

In center frequency estimation, we see that under ideal channel condition under which many commonly used estimators are derived, performance is very good for large block size and high SNR cases, offering a level of precision which we are currently unable to attain from similar learned ANN approaches.  However, we can see that at lower SNR levels, for smaller block sizes, and for harsh non-impulsive fading channel conditions the learned ANN based estimator approach appears to offer significant potential for improvement.  

The widespread use of learned approximate estimators is still in its infancy, such systems have not yet gained widespread acceptance or proven themselves to demonstrate all desirable properties under varying conditions, but we believe what we have shown here provides a valuable step towards this goal demonstrating that under certain constraints and requirements such systems do appear to outperform their baseline equivalents significantly based solely on regression of labeled datasets.

From a complexity standpoint we have shown that the ANN estimator complexity is within the same order of magnitude as the traditional estimators, is lower in some cases, and holds the potential to be significantly lower in complexity given additional optimization.

Going forward, approximate estimation based on regression of large datasets holds an extremely promising avenue of research when building practical engineering systems.  As we have shown in this paper, there are conditions such as short-time, and low-SNR conditions where accuracy gains can be achieved against current baselines, and there are also conditions where complexity reduction can be attained to reduce the computational complexity, size, weight, and power required in order to obtain a comparable estimate.  All of these properties must be traded in an engineering system design when selecting a design approach.  In future work we seek to further reduce the computational complexity through additional architecture optimization and further characterize the conditions under which each approach holds significant benefits or drawbacks versus the other.
\printbibliography

\end{document}